\documentclass{article}
\usepackage{arxiv_spconf,amsmath,graphicx}

\usepackage{multirow}
\usepackage{color}
\usepackage[hyphens]{url}
\usepackage[hidelinks]{hyperref}
\usepackage[hyphenbreaks]{breakurl}
\usepackage[capitalize]{cleveref}
\usepackage{enumitem}
\usepackage{etoolbox}
\usepackage{graphicx}
\usepackage{makecell}
\usepackage{caption}
\usepackage{wrapfig}
\usepackage{setspace}

\usepackage{floatflt}
\usepackage{float}

\graphicspath{ {./image/} }

\title{CONFORMER-BASED HYBRID ASR SYSTEM FOR SWITCHBOARD DATASET}
\name{Mohammad Zeineldeen$^{1,2*}$\thanks{$^*$Equal contribution}, Jingjing Xu$^{1*}$,
Christoph Lüscher$^{1,2}$, Wilfried Michel$^{1,2}$}
\secondlinename{Alexander Gerstenberger$^1$, Ralf Schlüter$^{1,2}$, Hermann Ney$^{1,2}$}
\address{
  $^1$Human Language Technology and Pattern Recognition, Computer Science Department, \\
  RWTH Aachen University, 52074 Aachen, Germany \\
  $^2$AppTek GmbH, 52062 Aachen, Germany
}

\renewrobustcmd{\bfseries}{\fontseries{b}\selectfont}
\renewrobustcmd{\boldmath}{}
\newrobustcmd{\B}{\bfseries}

\newsavebox\CBox

\makeatletter

\renewcommand{\section}{\@startsection
   {section}%
   {1}%
   {}%
   {-0.4\baselineskip}%
   {0.2\baselineskip}%
   {}}%

\renewcommand{\subsection}{\@startsection
  {subsection}%
  {2}%
  {}%
  {-0.1\baselineskip}%
  {0.1\baselineskip}%
  {}}%

\renewcommand{\subsubsection}{\@startsection
  {subsubsection}%
  {3}%
  {}%
  {-0.2\baselineskip}%
  {0.2\baselineskip}%
  {}}%

\g@addto@macro\normalsize{%
  \setlength\abovedisplayskip{3pt plus 2pt minus 1pt}
  \setlength\belowdisplayskip{3pt plus 2pt minus 1pt}
  \setlength\abovedisplayshortskip{2pt plus 2pt minus 1pt}
  \setlength\belowdisplayshortskip{2pt plus 2pt minus 1pt}
}

\setlist{
    itemsep=0pt,
    parsep=1pt plus 1pt minus 1pt,
    topsep=1pt plus 1pt minus 1pt,
    partopsep=0pt
}

\begin{document}

\maketitle
\begin{abstract}
The recently proposed conformer architecture has been successfully used
for end-to-end automatic speech recognition (ASR) architectures
achieving state-of-the-art performance on different datasets.
To our best knowledge, the impact of using conformer acoustic model for
hybrid ASR is not investigated.
In this paper, we present and evaluate a competitive conformer-based hybrid
model training recipe.
We study different training aspects and methods to improve word-error-rate
as well as to increase training speed.
We apply time downsampling methods for efficient training and
use transposed convolutions to upsample the output sequence again.
We conduct experiments on Switchboard 300h dataset and our conformer-based
hybrid model achieves competitive results compared to other architectures.
It generalizes very well on Hub5'01 test set and outperforms the BLSTM-based
hybrid model significantly.
\end{abstract}
\section{Introduction \& Related Work}
\label{sec:intro}

Hybrid neural network (NN)-hidden Markov model (HMM) automatic speech
recognition (ASR) systems \cite{bourlard2012connectionist}
have achieved state-of-the-art performance on different tasks
\cite{zhou2020rwth,Lscher2019RWTHAS,kitza2019cumulative}.
Bi-directional long short-term memory (BLSTM) \cite{hochreiter1997long}
has been widely used for acoustic modeling for conventional hybrid ASR systems.
Other neural architectures such as time-delay neural networks
\cite{peddinti2015time} and convolutional neural networks (CNN)
\cite{abdel2014convolutional} were studied.
The NN acoustic models (AMs) are often trained with cross-entropy using a
frame-wise alignment generated by a Gaussian mixture model (GMM)-HMM
system.

In the last few years, self-attention networks
\cite{Sperber2018SelfAttentionalAM}
have been shown to be better in terms of word-error-rate (WER)
as well as training speed since the self-attention mechanism
can be easily parallelized.
Transformer-based hybrid models \cite{Yong2019TransformerAM} have been
investigated and shown to be very competitive.
Recently, the conformer model \cite{gulati2020conformer},
which is another self-attention-based model, was proposed and achieved
state-of-the-art performance on Librispeech 960h dataset \cite{libri_dataset}.
It uses a convolution module to capture local context
dependencies in addition to the long context captured by the
self-attention module.
The conformer architecture was investigated for different
end-to-end systems such as attention encoder-decoder models
\cite{wang2021efficient,Zoltan2021conformer},
and recurrent neural network transducer
\cite{gulati2020conformer,burchi2021efficient}.
Nevertheless, there has been no work investigating the impact of
using a conformer AM for hybrid ASR systems.

In addition, the self-attention mechanism requires more memory resources and
also, the time complexity grows quadratically with the sequence length.
Thus, time downsampling techniques were introduced, mainly for end-to-end
systems, such as pyramidal reduction
\cite{Chan2016LAS}, max-pooling layers \cite{Zeyer_2018},
strided convolution \cite{han2020contextnet}, etc.
When training hybrid ASR models with frame-wise alignment,
upsampling the sequence at the output layer allows us to reuse the
existing alignments.
In this work, we utilize a simple but yet effective subsampling method by using
strided convolution for downsampling and a
transposed convolution \cite{long2015fully} for upsampling.
This method increases training speed and reduces memory consumption.
This helps to do sequence discriminative training efficiently.


In this work, we propose and evaluate a conformer-based hybrid ASR
system.
To our best knowledge, there is no such training recipe for
conformer-based hybrid models in the literature.
We explore different aspects to improve the WER as well
as training speed.
We apply time subsampling techniques using strided convolution for
downsampling and transposed convolution for upsampling.
We also investigate the effect of parameter sharing between
different components of the model.
In addition, we use a method, called LongSkip, to connect the
VGG \cite{simonyan2015deep} output directly to each conformer
block which lead to moderate improvement.
Moreover, due to time downsampling, it is possible
to do sequence discriminative training efficiently as it
reduces the memory usage requirements.
We present improvements on top of the BLSTM hybrid model and competitive
results on the Switchboard 300h Hub5'00 and Hub5'01 datasets.

\begin{figure}[t]
  \centering
  \includegraphics[width=8cm]{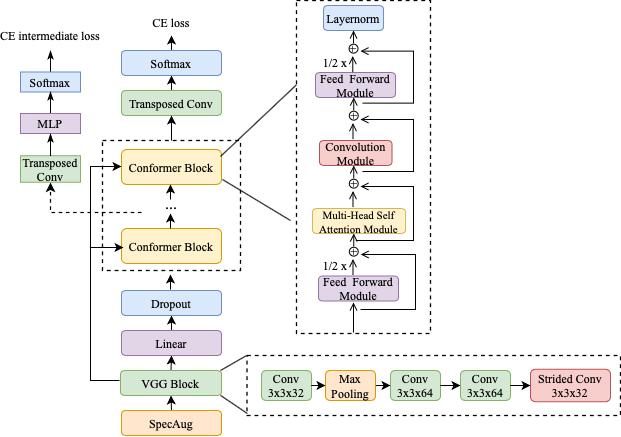}
  \caption{\textit{Overview of the proposed conformer-based hybrid model with
  time downsampling, and other training methods.
  See \Cref{sec:training_methods,subsec:baseline} for more description.}}
  \label{fig:conformer_arch}
\end{figure}

\section{Conformer Architecture}

The standard conformer architecture \cite{gulati2020conformer} consists
mainly of four modules:
feed-forward (FFN) module, multi-head \mbox{self-attention} (MHSA) module,
convolution (Conv) module, and another feed-forward module.
Let $x$ be the input sequence to conformer block $i$, then the equations
can be defined as below:
\[
  x_{FFN_1} = x + \dfrac{1}{2}\mathrm{FFN}(x)
\]
\[
  x_{MHSA} = x_{FFN_1} + \mathrm{MHSA}(x_{FFN_1})
\]
\[
  x_{Conv} = x_{MHSA} + \mathrm{Conv}(x_{MHSA})
\]
\[
  x_{FFN_2} = x_{Conv} + \dfrac{1}{2}\mathrm{FFN}(x_{Conv})
\]
\[
  \text{ConformerBlock}_i = \mathrm{LayerNorm}(x_{FFN_2})
\]

The illustration about the proposed conformer AM architecture can be
seen in \Cref{fig:conformer_arch}.

\section{Training Methods}
\label{sec:training_methods}

To build and improve the training recipe for the conformer-based
hybrid model, we investigate different training methods inspired
from the literature.
This helped to improve the WER as well as the training speed
and memory efficiency.

\textbf{Time down-/up-sampling:} The self-attention mechanism requires
allocating the whole input batch sequences into memory.
Due to memory constraints, this can lead to batch size reduction making
training slower.
To handle this issue, time downsampling techniques can be applied.
In this work, we use a strided convolution as part of the VGG network
for downsampling.
However, it is not straightforward to apply such downsampling methods
for models trained with frame-wise target alignment.
This will lead to a mismatch in the number of targets for computing the
frame-wise loss objective function.
To fix this issue, we use transposed convolution \cite{long2015fully} inspired
from computer vision in order to upsample again to the frame-wise target
alignment length.
For consistency, the filter size and the stride of the transposed convolution
are set to the time reduction factor.

\textbf{Intermediate loss:} Training deeper networks requires careful tricks
for convergence.
Adding intermediate or auxiliary losses
\cite{Yong2019TransformerAM,tjandra2020dejavu} at different layers
has been shown to be effective for training stability.
Note that when we do downsampling, a transposed convolution layer is
needed for upsampling to compute the intermediate loss.

\textbf{Parameter sharing} can be used to reduce model size.
We investigate the effect of sharing the parameters of the intermediate
loss layers as well as the ones of the transposed convolution layers.

\textbf{LongSkip} is a simple residual connection \cite{he2015deep}
variant similar to densely connected convolution neural networks
\cite{Huang2017DenselyCC}.
The motivation behind this is to connect a layer with previous layers to reuse
learned features.
In our case, we connect the output of the VGG network to the input of each
conformer block which leads to some improvement.

\textbf{Focal loss} \cite{lin2018focal} is a method that reshapes the CE
objective function to down-weight well-classified targets
and thus makes the model focus on misclassified targets.

\textbf{SpecAugment} \cite{Park_2019} is a data augmentation method
that masks out blocks of frequency channels and blocks of time steps.

\textbf{Sequence discriminative training} \cite{vesely2013sequence}
reduces the mismatch between training and recognition.
We use that to further improve the performance of the conformer hybrid model.

\section{Experimental Setup}
\label{sec:exp_setup}

All acoustic models are trained on Switchboard 300h dataset
\cite{Godfrey1992SWB} which consists of English telephony conversations.
We use Hub5'00 as development set which consists of Switchboard (SWB) and
CallHome (CH) parts.
We use Hub5'01 as test set.
We use RASR \cite{Wiesler2014RASR} for feature extraction and recognition.
RETURNN \cite{Zeyer_2018} is used to train the acoustic models.
All our config files and code to reproduce the results can be found
online\footnote{\scriptsize\url{https://github.com/rwth-i6/returnn-experiments/tree/master/2021-swb-conformer-hybrid}}.

\subsection{Baseline}
\label{subsec:baseline}

For NN training, we use 40-dimensional Gammatone features
\cite{schluter2007Gammatone}.
The first block of the NN consists of a VGG network similar to
\cite{simonyan2015deep}.
We use 4 convolution layers each having $3 \times 3$ kernel size.
The number of output filters for each layer are 32, 64, 64, 32 respectively.
We apply Swish activation \cite{ramachandran2017searching} between all
convolution layers which we observe to be better than using ReLU
\cite{agarap2018deep}.
Moreover, we apply max-pooling layer over feature dimension between first
and second convolution layers.
The last convolution layer is a strided convolution used for
time downsampling by factor of 3.
This is followed by 12 conformer blocks.
For time upsampling, a transposed convolution is used.
The attention dimension of each MHSA module is 512
with 8 attention heads.
The dimension of the feed-forward module is 2048.
We also use relative positional encoding.
%
The output labels consists of state-tied triphones using CART
\cite{Young1994CART}.
The number of CART labels is 9001.
We add two intermediate loss layers on the output of the $4^{th}$ and $8^{th}$
conformer blocks.
These layers consist of a transposed convolution for upsampling followed by an MLP
of one linear projection layer with dimension $512 \times 512$.
For training, we use frame-wise cross-entropy (CE) criteria.
The target frame-wise alignment is generated using a HMM-GMM system
from this setup \cite{zoltan_gmm}.
We use Adam optimizer with Nesterov momentum (Nadam)
\cite{dozat2016incorporating}.
Newbob learning schedule \cite{Zeyer_2017} with decay factor of 0.9 is applied
to control the learning rate based on CE development set scores.
We apply dropout of $10\%$ for all conformer modules as well as embedding and
attention dropout.
We apply weight decay \cite{NIPS1991_8eefcfdf} with a value of 0.01 to the
transposed convolution layers.
We observe that applying weight decay to other layers hurts the performance.
To avoid overfitting, we use focal loss \cite{lin2018focal} with a factor of 2.
We apply linear learning rate warmup from 0.0002 to 0.018 for 1.6 epochs.
We use batch size of 10k frames.
Batches are constructed with shuffled data based on sequence length
distribution.
The baseline is trained for 27 epochs.

\subsection{Language Models}
In recognition, we utilize 4-gram count-based language model (LM) and LSTM
LM as first pass decoding \cite{beck2019lstm}.
The LSTM LM has 51.3 perplexity (PPL) on Hub5'00 dataset.
We use a transformer (Trafo) LM for rescoring having PPL of 48.1 on Hub5'00.

\section{Experimental Results}

In this section, our training recipe for the conformer-based hybrid model
is investigated.
We conduct various experiments to understand the impact of each training
method and also to find the current optimum recipe.

\subsection{Depthwise Convolution Kernel Size}

In \Cref{tab:depthwise_conv_kernel_size}, we experiment with different
kernel sizes for the depthwise convolution.
We can observe that the kernel size has a significant effect on WER.
The model performs much better with smaller kernel size and the best WER
is achieved with kernel size 8.
This is also consistent with downsampling effect as the local context
is sampled.

\begin{table}[t]
  \centering
  \caption{\textit{WERs [\%] of using different kernel sizes for depthwise
  convolution. 4-gram count-based LM is used.}}
  \label{tab:depthwise_conv_kernel_size}
  \begin{tabular}{|c|c|c|c|}
    \hline
    \multirow{3}{*}{\shortstack{Kernel \\ size}} & \multicolumn{3}{c|}{WER [\%]} \\ \cline{2-4}
                                & \multicolumn{3}{c|}{Hub5'00} \\ \cline{2-4}
    & SWB & CH & Total \\ \hline
    6 & 8.4 & 17.1 & 12.8\\ \hline
    \textbf{8} & \textbf{8.1} & \textbf{16.8} & \textbf{12.5} \\ \hline
    16 & 8.2 & 17.6 & 12.9 \\ \hline
    32 & 8.4 & 18.0 & 13.2 \\ \hline
  \end{tabular}
\end{table}

\subsection{Number of Conformer Blocks}

We conduct experiments with different number of conformer blocks using the best
training recipe.
Results are shown in \Cref{tab:num_blocks}.
We can observe that we gain performance as we use deeper network.
Due to memory constraints, we use 12 conformer blocks as baseline for our
experiments.

\begin{table}[t]
    \centering
    \caption{\textit{WERs [\%] of using different number of conformer blocks.
    $\mathrm{L}$ is the number of conformer blocks.
    4-gram count-based LM is used.}}
    \label{tab:num_blocks}
     \begin{tabular}{|c|c|c|c|c|}
    \hline
    \multirow{3}{*}{$\mathrm{L}$}
    & \multirow{3}{*}{\shortstack{Params. \\ $[\text{M}]$}}
    & \multicolumn{3}{c|}{WER [\%]} \\ \cline{3-5}
    && \multicolumn{3}{c|}{Hub5'00} \\ \cline{3-5}
    && SWB & CH & Total \\ \hline
    6 & 42 & 8.5 & 18.0 & 13.3 \\ \hline
    8 & 59 & \textbf{8.1} & 17.3 & 12.7 \\ \hline
    12 & 88 & \textbf{8.1} & \textbf{16.8} & \textbf{12.5} \\ \hline
  \end{tabular}
\end{table}

\subsection{Time Downsampling Factors and Variants}

In \Cref{tab:time_downsampling_factor}, we report results with different
time downsampling factors showing tradeoff between speed and
\mbox{performance}.
We use downsampling by factor of 3 for the baseline as compromise.
In addition to that, \Cref{tab:time_downsampling_variants} shows results
with different variants of downsampling.
The BLSTM+maxpool variant consists of one BLSTM layer with 512
units in each direction followed by a time max-pooling layer.
VGG-layerX means that we use a strided convolution as the X$^{th}$ layer of the
VGG network.
Results show that strided convolution works better for downsampling.
It is also better to apply downsampling at the end of the VGG network which
allows it to encode more information.

\begin{table}[t]
    \centering
    \caption{\textit{WERs [\%] of applying different time downsampling factors.
    Training time is reported over 1/6 split of train data using a single
    GeForce GTX 1080 Ti GPU.
    4-gram count-based LM is used.}}
    \label{tab:time_downsampling_factor}
     \begin{tabular}{|c|c|c|c|c|}
    \hline
    \multirow{3}{*}{\shortstack{Factor}} &
    \multirow{3}{*}{\shortstack{Train \\ time [h]}} & \multicolumn{3}{c|}{WER [\%]} \\ \cline{3-5}
    && \multicolumn{3}{c|}{Hub5'00} \\ \cline{3-5}
    && SWB & CH & Total \\ \hline
    2 & 1.28 & 8.3 &  16.4    &  12.4 \\ \hline
    3 & 0.92 & 8.1  & 16.8 & 12.5 \\ \hline
    4 & 0.86 & 8.4  & 17.9 & 13.2 \\ \hline
    5 & 0.73 & 8.7  & 18.6 & 13.7 \\ \hline
  \end{tabular}
\end{table}

\begin{table}[t]
    \centering
    \caption{\textit{WERs [\%] for front-end and time downsampling variants.
    Downsampling factor of 3 is used. 4-gram count-based LM is used.}}
    \label{tab:time_downsampling_variants}
     \begin{tabular}{|c|c|c|c|}
    \hline
    \multirow{3}{*}{Method} & \multicolumn{3}{c|}{WER [\%]} \\ \cline{2-4}
                            & \multicolumn{3}{c|}{Hub5'00} \\ \cline{2-4}
    & SWB & CH & Total \\ \hline
    BLSTM+maxpool & 8.2  & 17.0 & 12.7 \\ \hline
    VGG-layer2 & 8.4  & 17.7 & 13.1 \\ \hline
    VGG-layer4 & \textbf{8.1} & \textbf{16.8} & \textbf{12.5} \\ \hline
  \end{tabular}
\end{table}

\subsection{Ablation Study of Training Methods}

To better understand the importance of each training method described
in \Cref{sec:training_methods}, we switch off (-) or switch on (+)
only one method each time without re-optimizing the model.
The results are shown in \Cref{tab:ablation_study} and they are sorted
descendingly based on absolute WER on Hub5'00.
SpecAugment is the most important method where it helps significantly
to avoid overfitting and yields $20\%$ relative improvement in WER.
Using intermediate loss is also important for better convergence, which achieves
$7\%$ relative improvement in WER.
Moreover, sharing parameters between transposed convolutions helps.
Other training methods seem to have marginal improvements.

\begin{table}[t]
  \centering
  \caption{\textit{WERs [\%] of ablation study on the best training recipe.
  4-gram count-based LM is used.}}
  \label{tab:ablation_study}
  \begin{tabular}{|l|c|c|c|}
    \hline
    \multirow{3}{*}{Training method} & \multicolumn{3}{c|}{WER [\%]} \\ \cline{2-4}
                            & \multicolumn{3}{c|}{Hub5'00} \\ \cline{2-4}
    & SWB & CH & Total \\ \hline
    Baseline & \textbf{8.1} & \textbf{16.8} & \textbf{12.5} \\ \hline \hline
    - SpecAugment  & 9.8 & 21.5 & 15.7 \\ \hline
    - Intermediate loss & 8.9 & 18.1 & 13.5 \\ \hline
    - Share transp. conv params. & 8.5 & 17.3 & 12.9 \\ \hline
    - LongSkip & \textbf{8.1} & 17.2 & 12.7 \\ \hline
    - Focal Loss & \textbf{8.1} & 17.0 & 12.6 \\ \hline
    + Share MLP params. & 8.2 & 16.9 & \textbf{12.5} \\ \hline
  \end{tabular}
\end{table}

\subsection{Comparison between Conformer and BLSTM AM}

In \Cref{tab:BLSTM_vs_conformer}, we compare the performance between using a
BLSTM or conformer as AM architecture.
The BLSTM-based model consists of 6 BLSTM layers following
a well-optimized setup as here \cite{kitza2019cumulative}.
SpecAugment is used.
We can observe that using a BLSTM-based model improves the WER as we increase
the number of parameters but still worse than a less parameterized
conformer-based model.
In this case, with nearly comparable number of parameters, the conformer model
outperforms the BLSTM model by around $9\%$ relative in terms of WER.

\begin{table}[t]
    \centering
    \caption{\textit{Comparison of WERs [\%] between BLSTM and conformer
    AM architectures.
    All models are trained for 27 epochs.
    4-gram count-based LM is used.}}
    \label{tab:BLSTM_vs_conformer}
    \begin{tabular}{|c|c|c|c|}
	\hline
	AM & \shortstack{LSTM\\dim.} & \shortstack{Params.\\$[\text{M}]$} & Hub5'00 \\ \hline
	\multirow{5}{*}{BLSTM} & 500\phantom{0}  & 41\phantom{0}  & 14.2 \\ \cline{2-3}
						          	 & 600\phantom{0}  & 57\phantom{0}  & 13.8 \\ \cline{2-3}
						             & 700\phantom{0}  & 76\phantom{0}  & 13.8 \\ \cline{2-3}
							           & 800\phantom{0}  & 96\phantom{0}  & 13.7 \\ \cline{2-3}
							           & 1000 & 146 & 13.3 \\ \hline \hline
    Conformer & - & 88\phantom{0}   & \textbf{12.5} \\ \hline
    \end{tabular}
\end{table}

\subsection{Sequence Discriminative Training (seq. train)}

We use the lattice-based version of state-level minimum Bayes risk (sMBR)
criterion \cite{Gibson2006HypothesisSF}.
The lattices are generated using the best conformer AM and a
bigram LM.
We observe that using a bigram LM is better than using a 4-gram LM.
A small constant learning rate with value 1e-5 is used.
At the final output layer, we use CE loss smoothing with a factor of 0.1.
The sMBR loss scale is set to 0.9.
Sequence discriminative training leads to $5\%$ relative improvement as shown in
\Cref{tab:overall_results}.
sMBR training requires the full input sequence without chunking.
Thus, it is important to note that due to downsampling, it was possible to
train with sMBR efficiently since the memory usage requirement is reduced and
therefore we can use larger batch sizes.

\section{Overall Results \& Discussion}

We summarize our results in \Cref{tab:overall_results} and compare with
different modeling approaches and architectures from the literature.
Overall, our conformer-based hybrid model yields competitive results.
Compared to BLSTM hybrid, the conformer model with 4-gram LM is $14.4\%$
relatively better in WER on Hub5'00.
It is also $8.5\%$ relatively better with LSTM LM.
Moreover, our conformer model outperforms a well-trained RNN-T model
with much fewer epochs.
The model is also on par with a well-optimized BLSTM attention system
\cite{tuske2020single} on Hub5'01 test set.
However, the conformer hybrid model is still behind the
state-of-the-art conformer attention-based system, yet it is trained with
much smaller number of epochs and this seems to matter a lot.
Note also that \cite{Zoltan2021conformer,tuske2020single} use cross-utterance
LM \cite{irie2019utt} during recognition which boosts their
\mbox{performance}.

\begin{table}
  \caption{\textit{Overall WER [\%] comparison with literature.}}
  \label{tab:overall_results}
  \centering
  \setlength\tabcolsep{1pt}
  \begin{tabular}{|@{\hskip1pt}c@{\hskip1pt}|@{\hskip1pt}c@{\hskip1pt}
    |@{\hskip2pt}c@{\hskip2pt}|@{\hskip1pt}c@{\hskip1pt}|@{\hskip1pt}c@{\hskip1pt}|
    @{\hskip1pt}c@{\hskip1pt}|@{\hskip1pt}c@{\hskip1pt}
    |@{\hskip1pt}c@{\hskip1pt}|}
    \hline
    \multirow{3}{*}{Work} & \multirow{3}{*}{\#Epochs} & \multirow{3}{*}{Approach}
    & \multirow{3}{*}{AM} & \multirow{3}{*}{LM} &  \multirow{3}{*}{\shortstack{seq.\\train}} &
    \multicolumn{2}{c|}{WER [\%]} \\
    \cline{7-8}
    & & & & & &  \makecell{Hub\\5'00} & \makecell{Hub\\5'01} \\ \hline
    \multirow{2}{*}{\cite{kitza2019cumulative}} & \multirow{2}{*}{-} & \multirow{2}{*}{Hybrid} & \multirow{2}{*}{LSTM} & 4-gram & \multirow{2}{*}{yes} & 13.9 & \multirow{2}{*}{-} \\
    \cline{5-5} \cline{7-7}
     &  &  &  & LSTM &  &  11.7 &  \\ \cline{7-8} \cline{5-5} \hline
    \multirow{2}{*}{\cite{zhou2021phoneme}}  & \multirow{2}{*}{100}
    & \multirow{2}{*}{RNN-T} & \multirow{2}{*}{LSTM} & LSTM
    & \multirow{2}{*}{no} & 11.5 & 11.5 \\\cline{5-5}
    \cline{7-8}
    & & & & Trafo & & 11.2 & 11.2 \\ \hline
    \cite{tuske2020single} & 250
    & LAS & LSTM & LSTM & no & 9.8 & 10.1 \\ \hline
    \multirow{3}{*}{\cite{Zoltan2021conformer}} & \multirow{3}{*}{250}
    & \multirow{3}{*}{LAS} & \multirow{3}{*}{Conf.} & - &
    \multirow{3}{*}{no} & 9.9 & 10.1 \\ \cline{5-5} \cline{7-8}
    & &  &  & LSTM &  & 8.6 & 8.5 \\ \cline{5-5} \cline{7-8}
    & &  &  & Trafo &  & 8.4 & 8.5 \\
    \hline \hline
    \multirow{5}{*}{ours} & \multirow{5}{*}{27} & \multirow{5}{*}{Hybrid}
    & \multirow{5}{*}{Conf.} & 4-gram
    & \multirow{2}{*}{no} & 12.5 & 12.1 \\  \cline{7-8} \cline{5-5}
    &  &  &  & LSTM &  &  11.3 & 10.5 \\ \cline{7-8}\cline{5-6}
    &  &  &  & 4-gram & \multirow{3}{*}{yes} &  11.9 & 11.4 \\ \cline{5-5} \cline{7-8}
    &  &  &  & LSTM &  &  10.7 & 10.1 \\ \cline{5-5} \cline{7-8}
    &  &  &  & Trafo &   & \textbf{10.3} & \textbf{9.7} \\ \hline

  \end{tabular}
\end{table}

\section{Conclusion}

In this work, for the first time a training recipe for a conformer-based
hybrid model is evaluated.
We combined different training methods from the literature that boosted
the word-error-rate.
We successfully applied time downsampling using strided convolution
to speedup training and used transposed convolution as a simple method
to upsample again.
We observe that SpecAugment and intermediate loss layers are necessary
to achieve good performance.
Sharing parameters between transposed convolution layers leads to moderate
improvement.
Our model generalizes very well on the Switchboard 300h test
set Hub5'01 and outperforms the BLSTM-based hybrid model significantly.
We believe further improvements are still possible if we do speed
perturbation, speaker adaptation, and longer training epochs.

\vspace{-0.13cm}
\begin{center}
  \textbf{Acknowledgement}
\end{center}
\vspace{-0.16cm}
\begin{floatingfigure}[l]{0.15\textwidth}
  \vspace{-4mm}
  \begin{center}
    \includegraphics[width=0.15\textwidth]{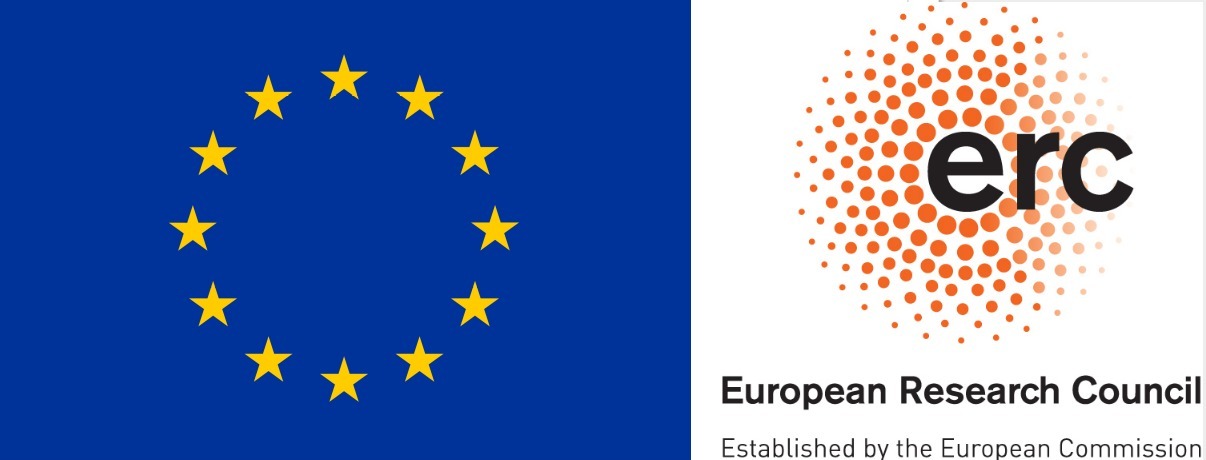} \\
  \end{center}
  \vspace{-3.5mm}
\end{floatingfigure}
\fontsize{8.2pt}{8.3pt}\selectfont
\setlength{\parindent}{0pt}
This project has received funding from the European Research Council (ERC)
under the European Union’s Horizon 2020 research and innovation programme
(grant agreement n\textsuperscript{o}~694537, project "SEQCLAS"). The work
reflects only the authors' views and the European Research Council
Executive Agency (ERCEA) is not responsible for any use that may be made of
the information it contains.
This work was partially supported by the project HYKIST funded by the
German Federal Ministry of Health on the basis of a decision of the
German Federal Parliament (Bundestag) under funding ID ZMVI1-2520DAT04A.

\vfill\pagebreak


\setstretch{0.6}

\renewcommand{\baselinestretch}{0.1}\normalsize
\let\OLDthebibliography\thebibliography
\renewcommand\thebibliography[1]{
  \OLDthebibliography{#1}
\let\normalsize\small\normalsize
  \setlength{\parskip}{0.8pt}
  \setlength{\itemsep}{1.1pt}
}

\bibliographystyle{IEEEbib}
\bibliography{refs}

\end{document}